# Building a Learning Database for the Neural Network Retrieval of Sea Surface Salinity from SMOS Brightness Temperatures

Adel Ammar, Sylvie Labroue, Estelle Obligis, Michel Crépon, and Sylvie Thiria

*Abstract* – This article deals with an important aspect of the neural network retrieval of sea surface salinity (SSS) from SMOS brightness temperatures (TBs). The neural network retrieval method is an empirical approach that offers the possibility of being independent from any theoretical emissivity model, during the in-flight phase. A Previous study [1] has proven that this approach is applicable to all pixels on ocean, by designing a set of neural networks with different inputs. The present study focuses on the choice of the learning database and demonstrates that a judicious distribution of the geophysical parameters allows to markedly reduce the systematic regional biases of the retrieved SSS, which are due to the high noise on the TBs. An equalization of the distribution of the geophysical parameters, followed by a new technique for boosting the learning process, makes the regional biases almost disappear for latitudes between 40°S and 40°N, while the global standard deviation remains between 0.6 psu (at the center of the of the swath) and 1 psu (at the edges).

## I. Introduction

ESA's Soil Moisture and Ocean Salinity (SMOS) mission [2], scheduled for launch in 2009, aims at providing the first global maps of sea surface salinity (*SSS*) and soil moisture from a series of brightness temperatures (*TBs*) measured at different incidence angles. We are interested in the inversion algorithm for the retrieval of the sea surface salinity (SSS) from the brightness temperatures (TBs), which is a critical issue for SMOS ground segment. We have chosen an empirical method based on neural networks (NN) as an alternative to the widely used iterative method [3] based on the inversion of a theoretical forward emissivity model method. The choice of an empirical method is justified by the imperfections of the existing theoretical models [4], [5]. Besides, the choice of neural networks is backed up by the large number of variables (several TBs at different incidence angles, and a set of geophysical parameters) and the non-linear dependence between them. Neural networks are non-linear statistical data modeling tools. They are used to model complex relationships between inputs and outputs, after a training phase over a learning database consisting of pairs of representative inputs and desired outputs.

A complete methodology of the neural network retrieval method, in the SMOS case, has been exposed in [1]. Section II synthesizes the main steps of this methodology. Besides, [1] has stressed the fact that the main drawback of this neural networks approach is that the high instrumental noise on TBs entails geographical biases on the retrieved SSS. Section III investigates the way to tackle this problem by working on the learning database of the inversion networks. Finally, section IV discusses the results obtained with the different databases that we have tested, and draws conclusions regarding the database that should be built in the operational phase.

## II. Methodology

One of the main difficulties of the neural network approach, in the SMOS case, is that measured *TBs* correspond to a set of incidence angles whose number and values vary from a surface point (pixel) to another, while the network inputs need to be fixed. This difficulty has been solved in [1], in two stages (Fig. 1). The pixels are first classified according to the range of incidence angles they are observed with (Fig. 2). Then we design a reasonable number of neural networks (10) having different ranges of inputs that correspond to the classes already determined. Finally, we apply an interpolation method to the available TBs at every surface point, to deduce the inputs of the corresponding network. The observed TBs are transformed into a series of TBs at fixed interpolated angles, corresponding to the inputs of the inversion networks. This is made using a locally weighted kernel linear estimation [6]. Then the pixels are processed through the suitable neural network.

The locally weighed regression also acts as a smoother, which is useful to reduce the high noise (due to the instrument and to the reconstruction process) on the TBs. Moreover, it allows dealing properly with unavailable measurements, through the adjustment of a variable bandwidth.

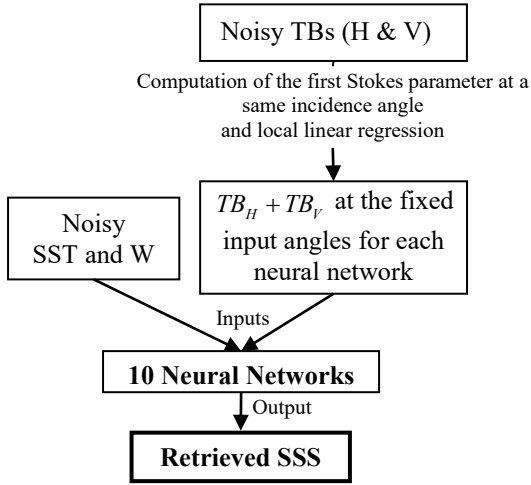

Fig. 1. Diagram of the SSS retrieval for a SMOS pixel.

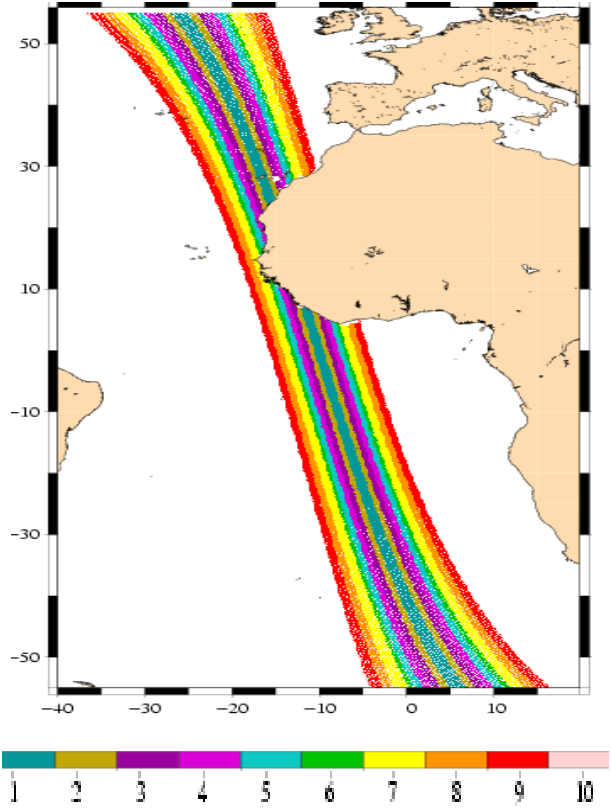

Fig. 2. Partition of the pixels observed during a SMOS half-orbit into 10 classes.

The inputs of the neural networks that were designed for the retrieval of *SSS* consist of a series of TBs, as well as the values of Sea Surface Temperature (*SST*) and Wind speed (*W*). Besides, we added a Gaussian random noise to *SST* (standard deviation of 1°C) and W (standard deviation of 2 m/s) to simulate the expected measuring errors on the ECMWF data that will be used in the operational phase. We do not need to add noise to SSS. The reference SSS is to be compared to the output of the neural networks to calculate the retrieval error.

We opted for the use of the first Stokes parameter (TBh +TBv) as an input of the retrieval algorithm, for the invariance property of this parameter when passing from the surface level to the antenna level [7], [8].

The radiometric noise has been simulated following [9] and [10]. Fig. 3 depicts the standard deviation of the simulated noise in a SMOS field of view. We simulated this noise independently for H and V polarizations and added it to the TBs resulting from SSA model [11].

The original noise is reduced after the interpolation phase that acts like a smoother [1]. For each class and each interpolation angle $Inc_0$, we computed the residual error after smoothing as follows:

$$\Delta T_{res}(Inc_0) = TB_{interpol}(Inc_0) - TB_{sim}(Inc_0)$$

Where $TB_{sim}$ is the "true" TB, simulated using SSA model (without adding noise). Table 1 shows the bias and standard deviation of the residual noise for each class. The standard deviation has been reduced by 30% to 60% depending on the class and the incidence angle, without introducing any significant bias.

Nevertheless, it is important to also take into account the correlation between the $\Delta T_{res}$ corresponding to neighbouring incidence angles. This correlation is induced by the interpolation phase, due to the overlapping of the bandwidths associated to neighbouring interpolation angles. Fig. 4 shows the values of the correlation matrix of the $\Delta T_{res}$ for the 23 interpolation angles of class n°1. The correlation matrix for class n°8 (3 interpolation angles) is the following:

$$\begin{pmatrix} 1 & 0.57 & -0.01 \\ 0.57 & 1 & 0.65 \\ -0.01 & 0.65 & 1 \end{pmatrix}$$

These correlations will be taken into account in section III when training the inversion networks using noisy TBs.

We used networks with only one hidden layer in which the number of hidden neurons is fixed empirically for each class. After fixing the architecture of the networks, they are trained using a learning database. Since we deal with simulated data, the global sea surface fields of the numerical ocean model MERCATOR (PSY3V1) have been used to form a learning database to train the neural networks, and then to evaluate the retrieval error on the global ocean. TBs are simulated using the SSA emissivity model [11].

For the in-flight processing, a new learning database will be built using *in situ* SSS data (provided by the ARGO network [12] collocated with SMOS observed TBs, in order to adjust the weights of the networks.

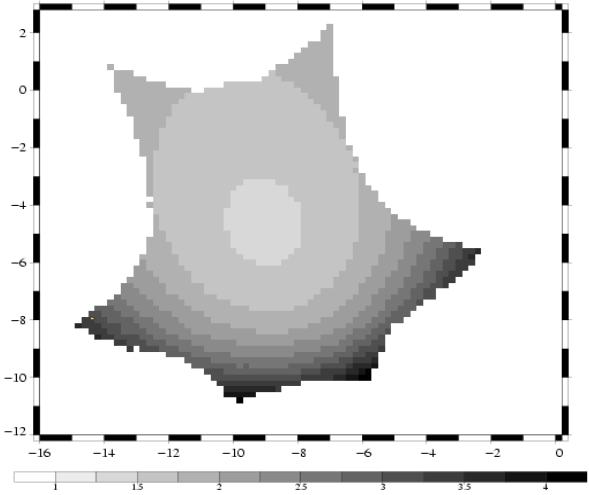

Fig. 3. Radiometric sensitivity calculated in a SMOS field of view (in K).

TABLE 1
RESIDUAL NOISE AFTER SMOOTHING

| Class / Network | Number of interpolated TBs | Distance across-track | Residual noise after smoothing (K) | |
|---|---|---|---|---|
| | | | Bias | Standard deviation (minimum and maximum over the interpolation angles) |
| 1 | 23 | 0-100 km | 0.015 | 0.9 – 1.1 |
| 2 | 21 | 100-150 km | 0.01 | 0.8 – 1.0 |
| 3 | 18 | 150-200 km | 0.015 | 0.8 – 1.1 |
| 4 | 16 | 200-250 km | 0.02 | 0.9 – 1.1 |
| 5 | 13 | 250-300 km | 0.012 | 1.0 – 1.1 |
| 6 | 10 | 300-330 km | 0.003 | 0.9 – 0.9 |
| 7 | 5 | 330-400 km | 0.019 | 1.0 – 1.2 |
| 8 | 3 | 400-470 km | -0.01 | 1.2 – 1.9 |
| 9 | 1 | 470-540 km | -0.0015 | 2.1 |
| 10 | 1 | 540-550 km | -0.048 | 1.9 |

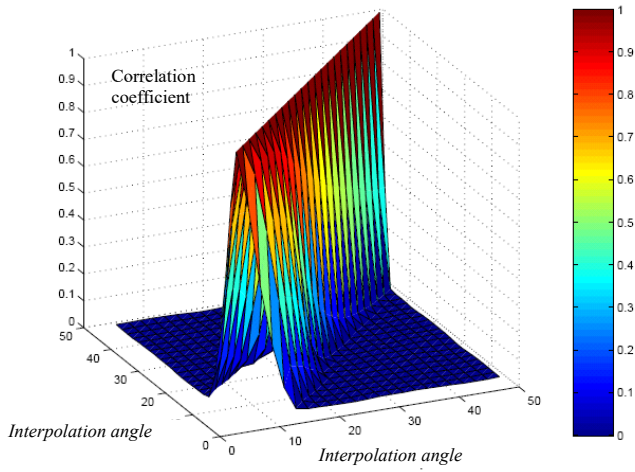

Fig. 4. Values of the correlation matrix of the residual noise on TBs (after smoothing), for class n°1.

## III. LEARNING DATABASE

The choice of the learning database is a critical issue, especially because of the high level of noise on the TBs. The quality of the learning database determines the quality of the SSS retrieval. In this section, we test different learning databases, and we evaluate their influence on the retrieval results.

We first built an initial database $B_0$ consisting of 12 MERCATOR PSY3V1 [13] daily global fields (one field per month, between September 2005 and August 2006, in order to take into account the seasonal variations), limited to latitudes between 65°S and 65°N, with a resolution of 0.5°. The different learning databases tested below are all extracted from $B_0$. This database will also be used to assess the retrieval performances, in order to have a global scale evaluation of the inversion algorithm. For computational reasons, the global maps were processed in their entirety with each of the 10 networks, independently of where the pixels would actually appear in the field of view.

Taking into account the results of the local linear interpolation, we added a random noise to the simulated *TBs* with the same standard deviation and correlation matrix as the residual noise $\Delta T_{res}$ (Table 1). In fact, the training must be conducted in the same conditions as when applied to realistic data.

### A. Learning Database $B_1$

The first learning database ($B_1$) used for the training of the neural networks resulted from a 2% random extraction from the initial database $B_0$. It consists of a set of 32,492 (SSS, SST, W) triplets. The random extraction provides a perfect representativeness of the geophysical situations on ocean. Fig. 5 shows the values of SSS in this learning database. We also use another independent database of 8123 triplets, built the same way, for cross-validation, in order to avoid overtraining.

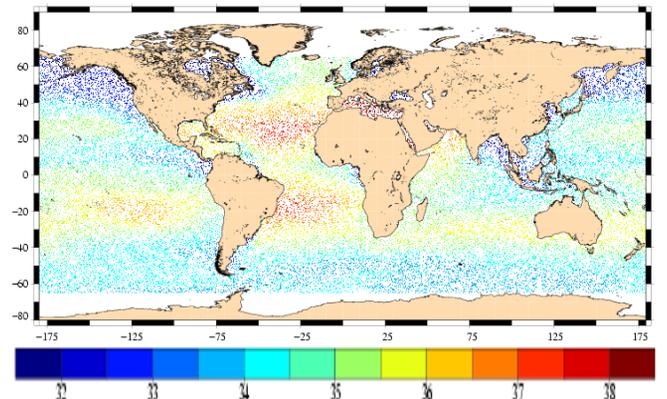

Fig. 5. SSS (psu) in learning database $B_1$.

After training with database $B_1$, we applied the network to the global test database $B_0$ (from which only 2% have been extracted to form the learning set). The retrieved SSS given by the output of each network is compared to the original SSS. The retrieval error is shown in Fig. 6 (class n°1) and Fig. 7 (class n°8).

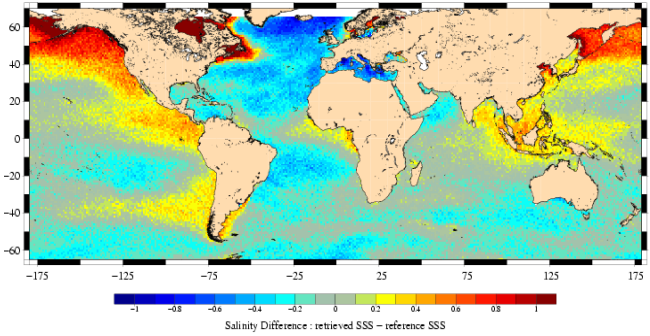

Fig. 6. 1°-sampling cartography of the retrieval mean error (in psu), for network n°1, on the global test database $B_0$, after using learning database $B_1$.

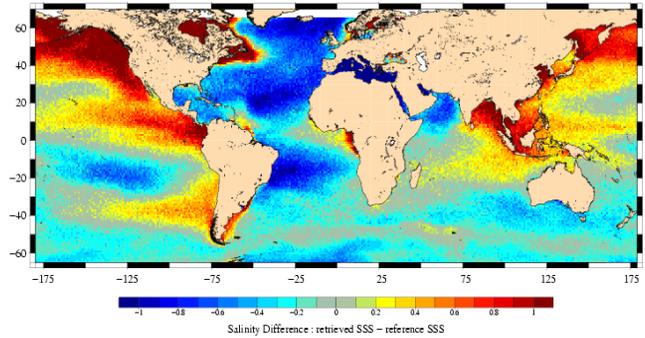

Fig. 7. 1°-sampling cartography of the retrieval mean error (in psu) for network n°8, on the global test database $B_0$, after using learning database $B_1$.

We obtained a global standard deviation of 0.50 psu for class n°1 and 0.66 psu for class n°8. There is no global bias, but low salinities (< 33 psu) are retrieved with a positive bias (0.76 psu, for class n°1), while high salinities (>35 psu) are retrieved with a negative bias (-0.16 psu, for class n°1). The slope of the regression line between retrieved and desired SSS is of 0.82 for class n°1 and 0.66 for class n°8. These figures point out the fact that high salinities are underestimated, while low salinities are overestimated. So the retrieval error is correlated to the SSS value. The correlation is more marked for the network n°8 because it has fewer inputs.

We do not present the results of all the ten classes, because they are similar in so far as they present systematic biases at the same regions. The values of the biases are however more important for classes with fewer inputs (class n°1 presents 23 input TBs, whereas class n°8 presents only 3 input TBs. See Table 1).

These regional biases are a serious problem, since SMOS level 2 products (retrieved SSS at the level of the pixel) are averaged spatially (100kmx100km) and temporally (30 days) to obtain level 3 products with a precision requirement of 0.1 psu to 0.2 psu [14]. This precision can not be met if the averaged errors are too correlated, which is the case here.

The systematic biases are due to the fact that the cost function of the inversion network is minimised according to the output SSS, while the significant noise is on the inputs TBs. Let's assume that $SSS = a.TB + b$, where $a$ and $b$ are constants, as described by the emissivity models, when SST and W are fixed (and for a given incidence angles). After adding an independent white noise $\eta$ to the TBs, the slope of the regression line between SSS and TB becomes:

$$a' = \frac{\text{cov}(TB+\eta, SSS)}{\text{var}(TB+\eta)} = \frac{\text{cov}(TB, SSS)}{\text{var}(TB) + \text{var}(\eta)} = \frac{a}{1 + \frac{\text{var}(\eta)}{\text{var}(TB)}}$$

(1)

So the addition of a white noise on the TBs distorts the statistical relationship between SSS and TB by reducing the slope of the regression line. The reduction factor increases with the ratio between the variance of the noise and the variance of TB.

### B. A Geophysically Equalized Learning Database ($B_2$)

Choosing a learning database that is representative of the geophysical parameters' distribution on ocean (as is the case of $B_1$) yields a bell-shaped histogram of SSS, with an important peak corresponding to mean salinities around 34 psu. Such a distribution increases the effects of the input noise: salinities situated on the left of the histogram peak are overestimated (because the noise on TBs makes them mixed up with mean salinities) whereas salinities situated on the right of the peak are underestimated (for the same reason). At fixed SST and W, a bell-shaped histogram of SSS yields a reduced TB variance (excluding noise), since this variance is proportional to the SSS variance. As a result, according to (1), the statistical relationship between SSS and noisy TBs, in the learning database, becomes highly distorted.

Consequently, in order to limit the noise effects, the SSS variance (at fixed SST and W) should be maximized in the learning database. A compromise between having a representative learning database and maximizing the SSS variance consists in opting for a constant density distribution of SSS for each situation of SST and W, which means an equalized distribution in the (SSS, SST, W) domain.

Therefore, we tested a learning database $B_2$ with a constant number (10) of triplets (SSS, SST, W) per interval of 0.2 psu, 0.5°C, and 1 m/s. It was obtained by extracting the triplets randomly from boxes containing more than 10 pixels, and duplicating pixels in the boxes containing less than 10, to represent equally frequent and rare geophysical situations. We obtain a total of 270,000 pixels. Fig. 8 depicts the geographical distribution of the pixels in database $B_2$. The regions having a high variability in terms of geophysical parameters are overrepresented.

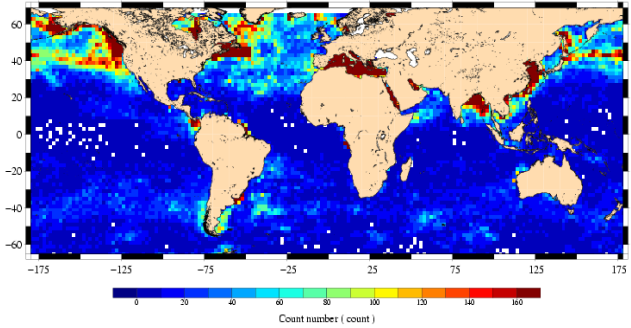

Fig. 8. Geographical distribution of the pixels in learning database $B_2$ (number of pixels in 2°x2° boxes).

After training with database $B_2$, we applied the network to the global test database $B_0$. The retrieval error is shown in Fig. 9 (class 1) and Fig. 10 (class 8). Compared respectively to Fig. 6 and Fig. 7 (obtained with the non equalized learning database $B_1$, and with the same conditions of noise), we notice that the regional biases are drastically reduced, except for the region under 40° South, because this particular region corresponds to a range of geophysical parameters (SSS between 33.5 psu and 35 psu and SST < 10°C) that were relatively more represented in the non equalized database. A more detailed explanation of the lingering bias in this region is given in section D.

The evolution of the slope of the regression line between original and retrieved SSS is a strong indication of the reduction of systematic biases. This slope rose from 0.81 to 1.03 for class n°1, and from 0.64 to 1.02 for class n°8.

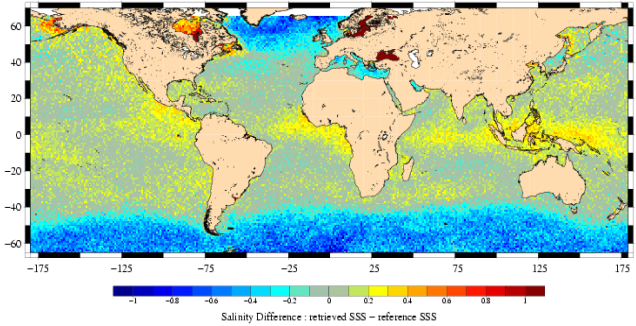

Fig. 9. 1°-sampling cartography of the retrieval mean error (in psu) for network n°1, on the global test database $B_0$, after using the equalized learning database $B_2$.

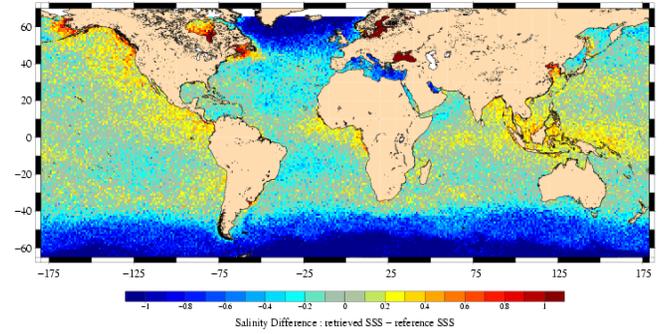

Fig. 10. 1°-sampling cartography of the retrieval mean error for network n°8, on the global test database $B_0$, after using the equalized learning database $B_2$.

### C. Boosting the learning process (database $B_3$)

In order to correct the systematic biases, we applied a new technique inspired by the boosting method used in classification problems [15]. The idea is to continue the training process only on a part of the learning database consisting of the poorly retrieved situations.

From the previous equalized learning database $B_2$, we extract the (SSS, SST, W) boxes (each one containing 10 pixels) that present a retrieval bias higher than a fixed threshold (0.2 psu). We obtain a new database $B_3$ containing 153,700 pixels (57% of $B_2$) that we use to extend the training process. We do not reinitialize the network weights. We reuse the same weights that have been obtained after the training on $B_2$, not to deteriorate the situations that have been correctly retrieved (with no important bias) in the previous training.

Fig. 11 shows the results obtained for network 1. Compared to Fig. 9, we notice that the systematic biases have almost disappeared in a large range of latitudes between 40°S and 40°N.

The extraction of database $B_3$ from database $B_2$ increases the weight of the poorly retrieved situations, which are generally at the edges of the learning domain in the (SSS, SST, W) space. This operation further increases the variance of SSS (for fixed SST and W), which reduces the noise effect, according to (1).

The regional biases in the southern latitudes have not disappeared, however. So another complementary solution should be considered for this particular region.

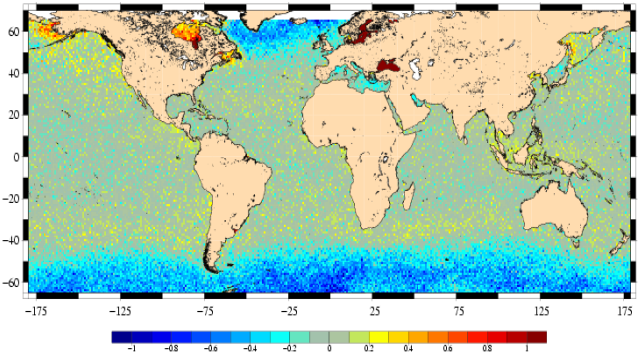

Fig. 11. 1°-sampling cartography of the retrieval mean error (in psu) for network n°1, on the global test database $B_0$, after a training on database $B_2$, followed by a training on database $B_3$.

### D. Correcting the Systematic Bias in the Southern Latitudes

The retrieval of the southern latitudes (under 45°S) turns out difficult. In fact, the equalization of the learning database according to the geophysical parameters (section III.B) induces a deterioration of the retrieval of this zone by introducing a systematic underestimation. The training boosting (section III.C) does not correct this drawback. This is due to the conjunction of two reasons:

1) The SSS variability in this zone is very low (0.3 psu in standard deviation). Consequently, the TB variance, at fixed SST and W, is also reduced. For mean values of SST and W (6±1°C and 9±1m/s), the standard deviation of TB (at nadir) is about 0.09 K, which is extremely low compared to the radiometric noise. So, according to (Eq. 1), the statistical connection between SSS and TB becomes highly distorted, which results in a systematic bias.
2) This region corresponds to low SST values (6°C on average), which means a lower sensitivity of TB to SSS [16].

In learning database $B_1$, the pixels of this region represent 82% of cold waters (SST<10°C), which counterbalances the two above-mentioned effects, and explains why this region is better retrieved than the high latitudes. On the contrary, in databases $B_2$ and $B_3$, the pixels of low latitudes represent only 41% and 44% (respectively) of cold waters. Extracting poorly retrieved situations (section III.C) does not significantly raise the weight of low latitudes in the learning database ($B_3$) because this extraction is made in the geophysical space (SSS, SST, W) where the low latitudes correspond to a relatively limited domain (especially because of the low SSS variability).

We put forward two distinct solutions to better the SSS retrieval in this particular region, either upstream by working on the distribution of the learning database, or downstream by combining the outputs of two inversion networks.

The first solution consists in building a mixed learning database $B_m$, where the distribution of the pixels is geographically regular (as in $B_1$) for cold waters (SST ≤ 10°C) and geophysically equalized (as in $B_2$) for warm waters (SST ≥ 10°C). We built such a database containing 270,000 pixels (same size as $B_2$). Fig. 12 shows the result obtained on test database $B_0$ after training the network of class n°1 with database $B_m$. Latitudes lower than 45°S are better retrieved compared to the result obtained with a totally equalized learning database ($B_2$). We obtain, in this region, a similar result to what we have obtained with learning database $B_1$. Nevertheless, the high latitudes (>40°N) show deteriorated results (compared to Fig. 9). If we extend the training (as in III.C), the retrieval of high latitudes is improved to the detriment of low latitudes, because the latter region is represented by a smaller domain in the (SSS, SST, W) space.

As for the second solution, it consists in combining the outputs of two inversion networks, each one trained on a different learning database. For instance, Fig. 13 shows the result obtained on test database $B_0$, after the following operation:
- For latitudes higher than 45°S, we select the output of the network (of class n°1) trained on database $B_1$.
- For latitudes lower than 50°S, we select the output of the network (of class n°1) trained on database $B_3$.
- For latitudes between 45°S and 50°S, we calculate a linear combination between the outputs of the above-mentioned networks, according to the latitude value. This is made to avoid edge effects.

We notice that the retrieval of low latitudes have been significantly improved compared to Fig. 13 (obtained with learning database $B_3$). A transition effect is still visible, however. To correct it, a more sophisticated combination, taking also into account the geophysical parameters could be investigated.

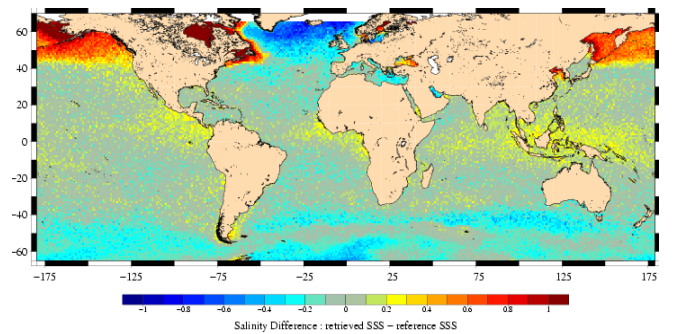

Fig. 12. 1°-sampling cartography of the retrieval mean error (in psu) for network n°1, on the global test database $B_0$, after using the mixed learning database $B_m$.

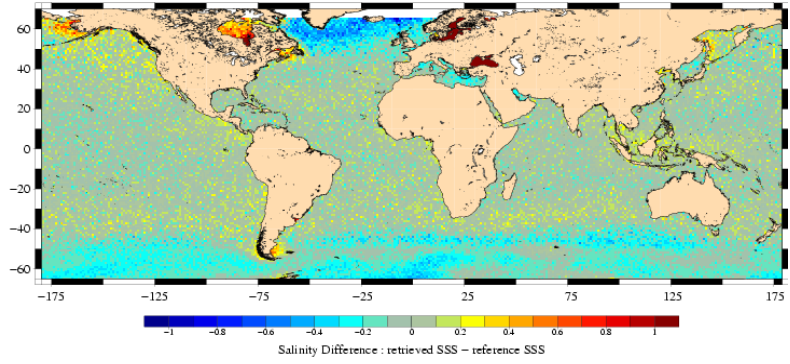

Fig. 13. 1°-sampling cartography of the retrieval mean error (in psu) for network n°1, on the global test database $B_0$, when combining the outputs of two networks trained respectively on $B_1$ and $B_3$.

TABLE 2
COMPARISON OF THE GLOBAL PERFORMANCES ON TEST DATABASE $B_0$, FOR CLASS N°1, AFTER TRAINING ON DIFFERENT DATABASES

|  | After training on B1 | After training on B2 | After training on B2, then on B3 | After training on Bm | Combination of the outputs of the networks trained on BA1 and BA3 |
|---|---|---|---|---|---|
| Global bias (psu) | 0 | -0.06 | -0.08 | 0 | -0.03 |
| Standard deviation (psu) | 0.50 | 0.64 | 0.67 | 0.53 | 0.58 |
| Slope of the regression line between retrieved and reference SSS | 0.81 | 1.03 | 1.02 | 0.90 | 0.99 |
| Percentage of 1°x1° boxes with a bias < -0.2 psu | 21% | 25% | 21% | 12% | 10% |
| Percentage of 1°x1° boxes with a bias > 0.2 psu | 19% | 10% | 3% | 9% | 4% |
| Percentage of 1°x1° boxes with a bias < -0.2 psu, for latitudes between 45°S and 65°N | 16% | 5% | 4% | 6% | 4% |
| Percentage of 1°x1° boxes with a bias > 0.2 psu, for latitudes between 45°S and 65°N | 19% | 10% | 3% | 9% | 3% |

## IV. DISCUSSION

Table 2 compares the retrieval performances on test database $B_0$, for class n°1 (pixels corresponding to the center of the swath), after a training on the different learning databases tested in section III. The minimum global bias and standard deviation are obtained with learning database $B_1$, since it is perfectly representative of the geophysical parameters' distribution in the test database $B_0$. In fact, the neural networks give more importance to situations that are the most represented in the learning database. Consequently, they are better retrieved. On the contrary, the situations that are relatively less represented in learning database $B_1$, are retrieved with an important bias: 50% of the 1°x1° geographical boxes (which contain 48 pixels, at the maximum) present a bias higher than 0.2 psu (in absolute value).

Learning database $B_2$ gives an equal weight to all geophysical situations. Consequently, the global standard deviation is higher (0.64 psu, for class n°1), since the frequent geophysical situations in $B_0$ are relatively less represented in $B_2$. Nevertheless, the regional biases are strongly reduced (apart from latitudes under 45°S). Training the inversion networks with database $B_3$ reduces further these biases while slightly increasing the global standard deviation. 93% of the 1°x1° geographical boxes, excluding latitudes under 45°S, present a bias lower than 0.2 psu (in absolute value).

The global slope between retrieved and reference SSS does not reflect the significant improvement obtained when training with database $B_3$. Nevertheless, when we calculate this slope per SST interval (Fig. 14), we notice a marked improvement for all temperatures. The slope remains however lower than 0.6 for cold waters (SST<5°C), because of the very poor sensitivity of TBs to SSS.

Learning database $B_m$ presents a compromise between databases $B_1$ and $B_2$, which allows a good retrieval of low latitudes and a low global standard deviation, but inherits the drawbacks of learning database $B_1$, as for the retrieval of high latitudes.

Finally, the combination of the outputs of two inversion networks trained on databases $B_1$ and $B_3$ respectively presents a simple solution that yields the lower biases for all latitudes and corrects the slope between retrieved and reference SSS for cold waters (Fig. 14).

The learning databases that we tested in this study will help to enhance the retrieval accuracy of the inversion algorithms during the in-flight processing, by choosing a judicious database based on in situ SSS data (provided by the ARGO network) collocated with SMOS observed TBs. The distribution of geophysical parameters in this new database should then be equalized, in order to minimize the systematic SSS biases, as has been shown in section III.B. The training process should then be conducted in two steps as explained in section III.C, while the particular zone of low latitudes may be processed separately.

As for the size of the learning database, it does not seem to be as much critical. For instance, we have tested a reduced learning database built the same way as $B_2$, but containing only 8000 pixels. This reduced database gave yet almost identical results (0.65 psu in standard deviation and a 1.02 slope between retrieved and reference SSS, for class n°1). Nevertheless, the size effect should be verified, in the operational phase, using SMOS observed TBs.

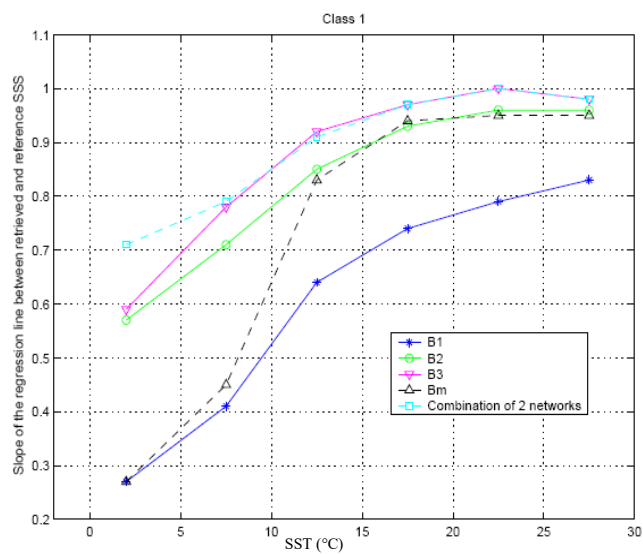

Fig. 14. Slope to the regression line between retrieved and reference SSS, calculated per interval of SST (<5°C, [5°C ;10°C], [10°C ;15°C], [15°C ;20°C], [20°C ;25°C], >25°C), after training with different databases.